Khithov, V., Petrov, A., Tishchenko, I., Yakovlev, K. (2016). Toward autonomous UAV landing based on infrared beacons and particle filtering. In Proceedings of *The 4th International Conference on Robot Intelligence Technology and Applications (RiTA 2015), Bucheon, Korea, December 14-16, 2015*. pp. 529-537. Springer International Publishing.

# Towards Autonomous UAV Landing Based on Infrared Beacons and Particle Filtering

Vsevolod Khithov, Alexander Petrov, Igor Tishchenko and Konstantin Yakovlev

Soloviev Rybinsk State Aviation Technical University, Rybinsk, Russia,
vskhitkov@gmail.com
NPP SATEK plus, Rybinsk, Russia
petrov@nppsatek.ru
Program System Institute of Russian Academy of Sciences, Pereslavl-Zalessky, Russia
igor.p.tishchenko@gmail.com
Institute For Systems Analysis of Russian Academy of Sciences, Moscow, Russia
yakovlev@isa.ru

**Abstract. Autonomous fixed-wing UAV landing based on differential GPS is now a mainstream providing reliable and precise landing. But the task still remains challenging when GPS aviability is limited like for military UAV's. We discuss a solution of this problem based on computer vision and dot markings along stationary or makeshift runway. We focus our attempts on using infrared beacons along with narrow-band filter as promising way to mark any makeshift runway and utilize particle filtering to fuse both IMU and visual data. We believe that unlike many other vision-based methods this solution is capable of tracking UAV position up to engines stop. System overview, algorithm description and it's evaluation on synthesized sequence along real recorded trajectory is presented.**

**Keywords: UAV, fixed-wing, autonomous landing, computer vision, particle filter, infrared markers, pattern detection, real-time, navigation, sensor fusing.**

## 1 Introduction

Unmanned Aerial Vehicles (UAV's) are widely used for surveillance, aerial photography and reconnaissance; payload carriers and military striking drones gain their popularity. Fixed-wing UAV's are known to have far longer flight times and distances than copter-like ones, but due to necessity of high speed maintenance during all mission to keep lift their landing is more challenging task. For fixed-wing UAV resque it's common practice to use parachute or catch them with net though disadvantages and limitations of such approach are obvious. Autonomous runway landing is commonly based on differential GPS. However, GPS can be not very reliable especially for military drones. Anyway, autonomous landing means real-time and precise localization of UAV pose relative to landing site along with keeping trajectory required for safe landing. While latter is a problem of control system, a lot of research is faced to obtaining drone's pose above runway using plenty of sensors including IMU, altimeter, camera, laser sensors, etc. under hard real-time constraints due to high relative speed of UAV. The situation can also

Authors' version as submitted to RiTA 2015. Publisher version is accessible at http://link.springer.com/chapter/10.1007/978-3-319-31293-4_43

be compounded by demand of reliable action under different weather conditions, day and night.

Our method refers to applications where tracking should be performed until a plane can switch to dead-reckoning by sensors on its landing gear, and internal navigation system is capable of initially guiding a plane to the beginning of landing approach with precision of about a hundred meters, knowing it's yaw, pitch and roll with accuracy of several degrees due to IMU and magnetic sensor. We also assume runway as planar.

### 1.1 Related works

There exist a lot of methods that use computer vision for runway detection by its edges or corners ([1], [2], [3] and many others). Such an approach is only suitable if some sort of asphalt concrete runway is used that has sharp boundaries or lane markings but not suitable for dirt runways.

Template search by correlation and edge-based tracking of landing site template image is utilized in [4], while in [5] mutual information criterion is considered for tracking. Such methods benefit from their versatility, but depend on landing site actual template availability and its parts' uniqueness. In [6], [7] SIFT features are used for matching between photo with known landing site position and video stream. All these methods suffer from scale problem – tracking is lost at too low altitude. In [8] scale problem is solved by using concentric circles marker as landing site for multirotor UAV – when outer circles are not longer visible, inner ones remain. The idea can hardly be generalized for large fixed-wing drones.

In [9] with the help of single near-infrared beacon center of a landing deck is determined, importance of correct camera exposure managing is shown. Near-infrared technology have already reached hobby UAV's [10], though the effective range is rather small. In [11], one of the early works, thermal imaging camera is used to detect "hot spots" of aircraft carrier to guide UAV along glidescope until low-power communication is allowed. Usage of mid-infrared (heated) beacons on a stationary runway is also discussed. Paper does not cover all aspects of autonomous landing, but usage of thermal range is promising. Thermal imaging is also used by creators of TAMS system [12], but unlike works above, infrared stereo camera and processing unit is placed on the ground and obtained UAV pose is delivered to drone's control system by means of existing communication channel. It's worth nothing that existing commertial landing systems (not using computer vision) like OPATS[13], UCARS[14] also have "active" ground equipment.

## 2 System overview

We discuss autonomous landing method based on visual servoing where infrared beacons placed along runway (their images) are used as features for UAV pose recovery. Light emitters are selected for practical reasons: first, some light emitting is anyway necessary if we want to achieve operation in low light conditions using camera; Second, such beacons can be quickly deployed on a temporary runway; third, distance of beacons' visibility is limited primarily not by their size but by intensity, that result in 1) easier solving of scale problem on different distances 2) extremely simple detection algorithm by threshold if we have enough light source power and manageable camera exposure. In practice, however, the method is not so easy, and challenges are: limited power of beacons, background illumination and simple shape of hotspots provide a lot of clutter to cope with; need to distinguish between identically looking beacon images.

System overview is illustrated on fig. 1. UAV is equipped with frontal-view camera with narrow-band filter, IMU, magnetic sensor and altimeter, all synchronized. It's supposed that in GPS-denied environment UAV's navigation system is still capable of guiding UAV to the vicinity of runway. We use three coordinate frames – for world, UAV and camera respectively. Beacons are placed in straight lines with regular distances along both sides of a runway. Straight lines and regular structure is chosen for reason of simple deploying.

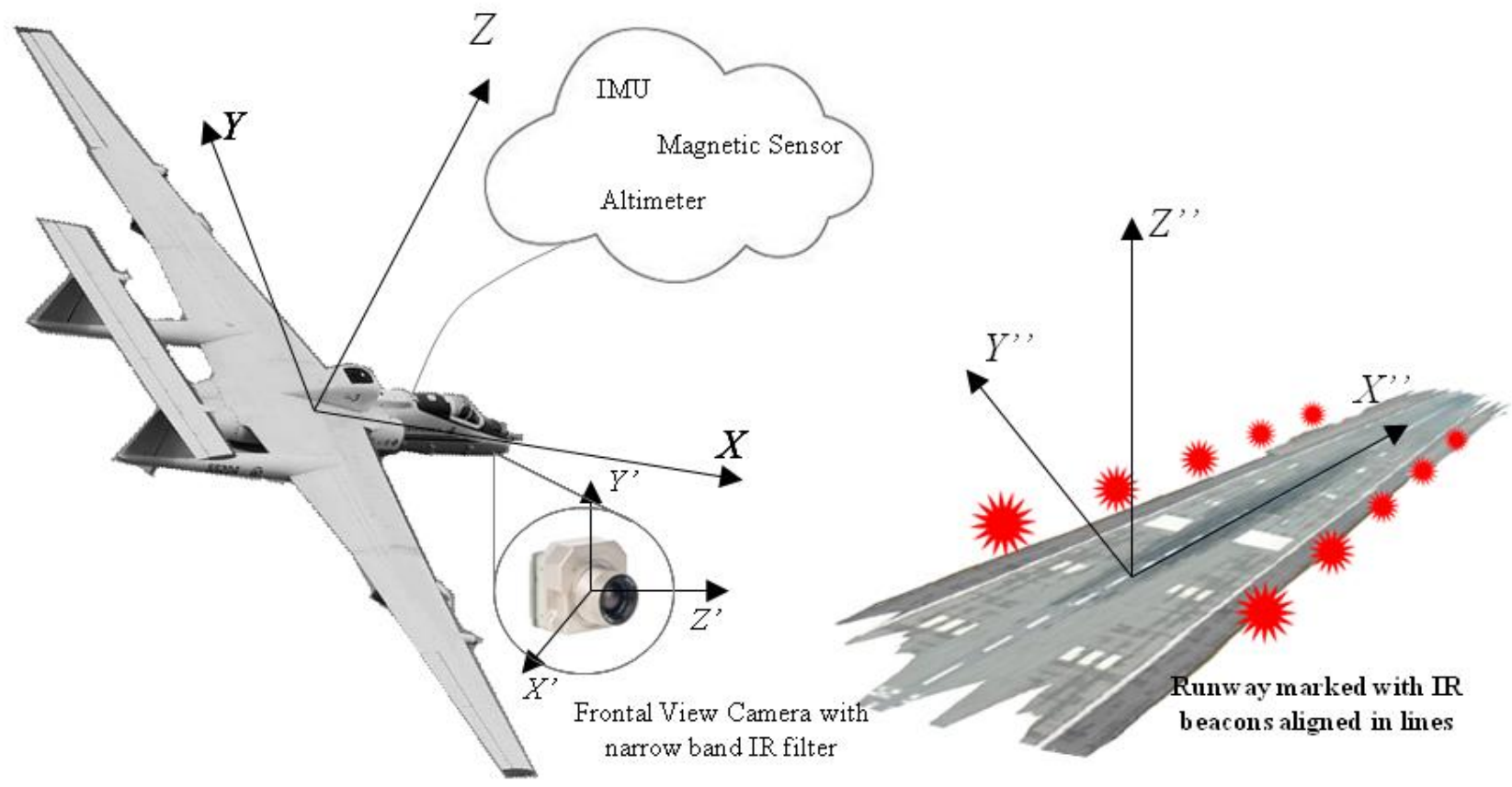

**Fig. 1.** Overview of landing system: UAV equipment, beacons configuration and coordinate frames

Camera position with respect to UAV and its intrinsic are ought to be known.

We look forward using MIR range (3-8 um) for beacon detection and tracking: mid-infrared beacons are just objects heated to several hundreds degrees so they can be very cheap while sun is emitting far less in this range than in near infrared. Narrow-band filters for the range also exist. However, thermal sensor technology is far more expensive then NIR, so we tried both. On fig. 2 (left) appearance of 30W NIR lamp through dielectric IR filter from 350 m distance on sunny day is shown. On fig. 2 (right) 20 cm object heated by portable gas burner is visible from 700 m distance by thermal camera.

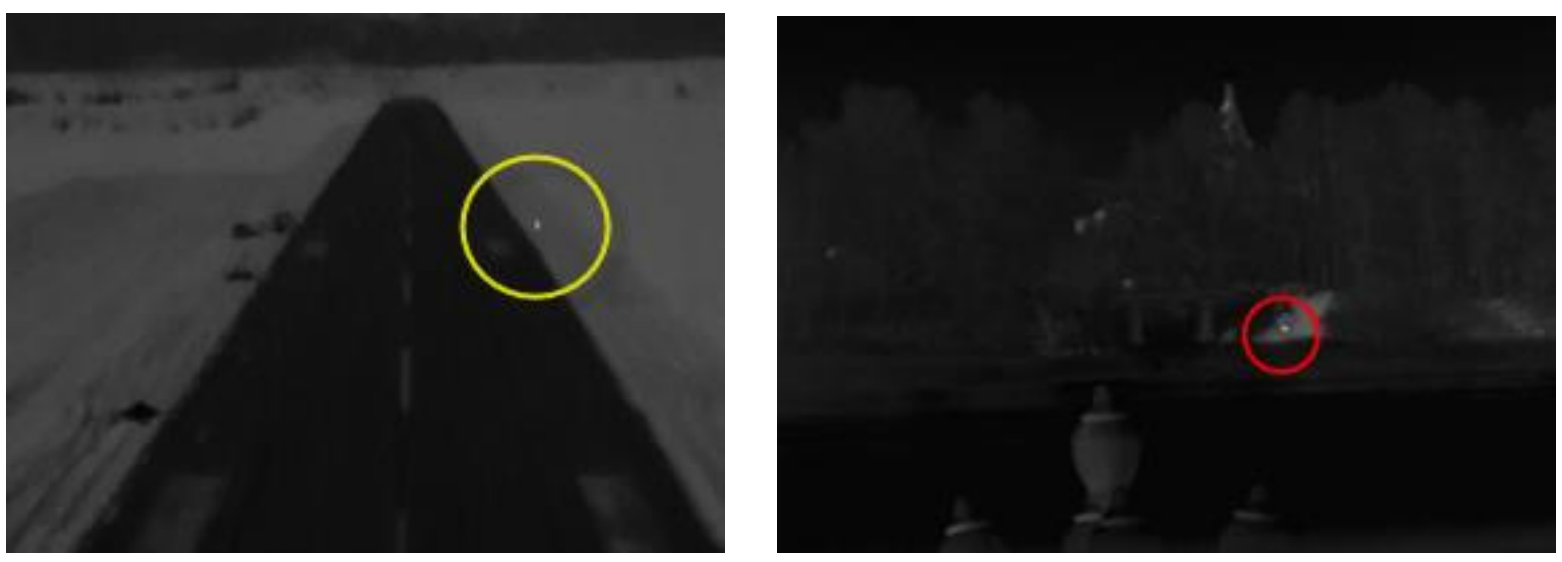

**Fig. 2.** *Left:* 30W NIR led lamp near runway visible through narrow-band IR filter in sunny day from 350 m; *Right:* 20 cm object heated by portable gas burner is visible from 700 m distance on thermal camera image in daylight conditions

## 3 Algorithm description

Particle filter is used to track UAV position relative to runway during descent. Let matrix $RT_{cam}$ be camera 3D pose in UAV frame in homogenous coordinates, $RT_{uav}$ - UAV 3D pose in world frame (unknown), $K$ – camera intrinsic matrix, $p^0{}_i$ – i-th beacon known position on the ground plane. Then we can obtain homography matrix that projects beacons' position $p_i$ from ground plane to the image plane using the following equation (1).

$$p_i = H \times p^0{}_i;$$
$$H = K \times \begin{bmatrix} 1 & 0 & 0 & 0 \\ 0 & 1 & 0 & 0 \\ 0 & 0 & 0 & 1 \end{bmatrix} \times RT_{cam}{}^{-1} \times RT_{uav}{}^{-1} \times \begin{bmatrix} 1 & 0 & 0 \\ 0 & 1 & 0 \\ 0 & 0 & 0 \\ 0 & 0 & 1 \end{bmatrix} \quad \textbf{(1)}$$

So, particle filter is used to model probability of UAV pose to be $RT_{uav}$ in the world frame. Generally there are six degrees of freedom, but pitch and roll angles are known with high accuracy from IMU synchronized with camera. Final state vector is ($\overline{T}, \gamma, \theta, \varphi, s$) where $\overline{T}$ is not a degree of freedom but denotes translation vector of UAV in world frame, $\gamma$ denotes azimuth, angles $\theta, \varphi$ are UAV movement direction azimuth and pitch respectively, $s$ is movement speed.

**Initialization**. Particle filter is initialized with probability distribution known from UAV's internal navigation system with typical uncertainty of some hundred meters in planar coordinates (GPS-denied action), ten meters in elevation and several degrees in azimuth.

**Prediction**. At each prediction step difference between previous and current IMU yaw measurement is added with slight Gaussian variance. $\overline{T}$ is changed according movement direction and speed (2)

$$\overline{T}^{(t)} = \overline{T}^{(t-1)} + R(\theta, \varphi) \cdot s \cdot \Delta t, \quad \textbf{(2)}$$

where $R(\theta, \varphi)$ is rotation matrix, $\Delta t$ – time increment.

**Measurement.** Measurement lies in matching projected beacons' positions for each hypothesis with incoming image. Matching consists of three stages:

- extracting light source binary map $M$ ;
- applying distance transform $D$ to binary map;
- each particle's weight calculation based on a equation upon distance map pixels' values $D_{p[i]}$ under projected beacons' positions $p_i$ .

When detecting light source, one should reasonably try to make as much contrast to background as possible while keeping dynamic range. This can be achieved by selecting appropriate exposure [9]. The law to choose exposure on far distances is to have as many as possible saturated pixels but do not have connected components of saturated pixels with area more than small predefined value. Moreover, as far as we now beacons' emission power and camera sensitivity, there is a range of automatic tuning of this parameter depending on how close to landing site UAV is.

The following algorithm is used for light source detection (Fig. 3), which resembles [15]. We model faint light sources as local maxima of incoming gray-scale image pixel intensities that are above k $k \approx 2.5$ standard deviations in this area.

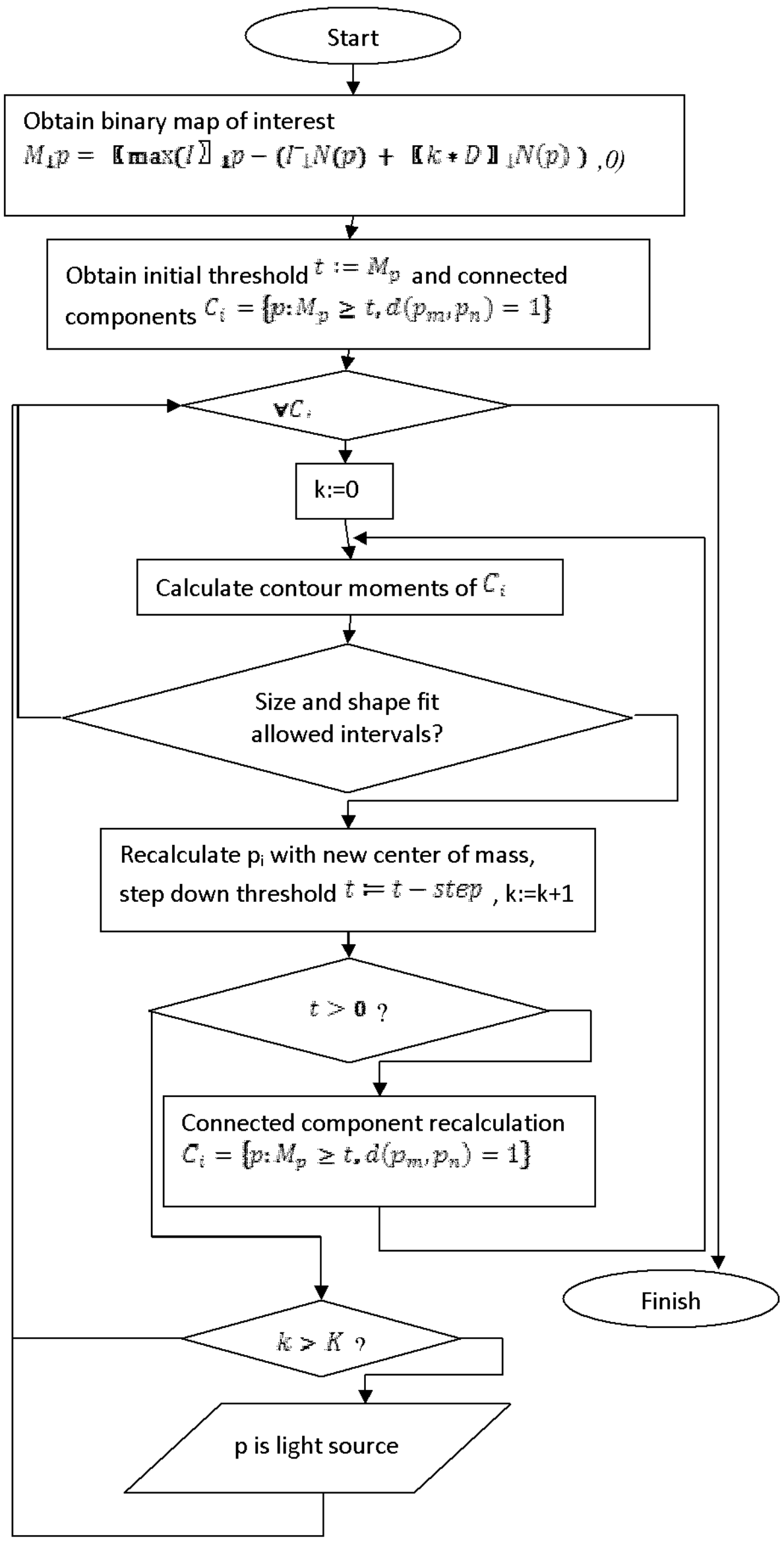

**Fig. 3**. Algorithm of light source detection. $I_p$ – source image, $I_{m \in N}$ - vicinity N of image, $D(I_{m \in N})$ – standard deviation of pixel intensities in vicinity.

We assign weight to a hypothesis with the following equation (3):

$$w = \exp\left(\sum_{i=1}^{N} \frac{D_{p[i]}}{\left(D_{p[i]} + P\right)^{q}}\right),$$

where q, P are parameters.
The pipeline is illustrated on fig. 4.

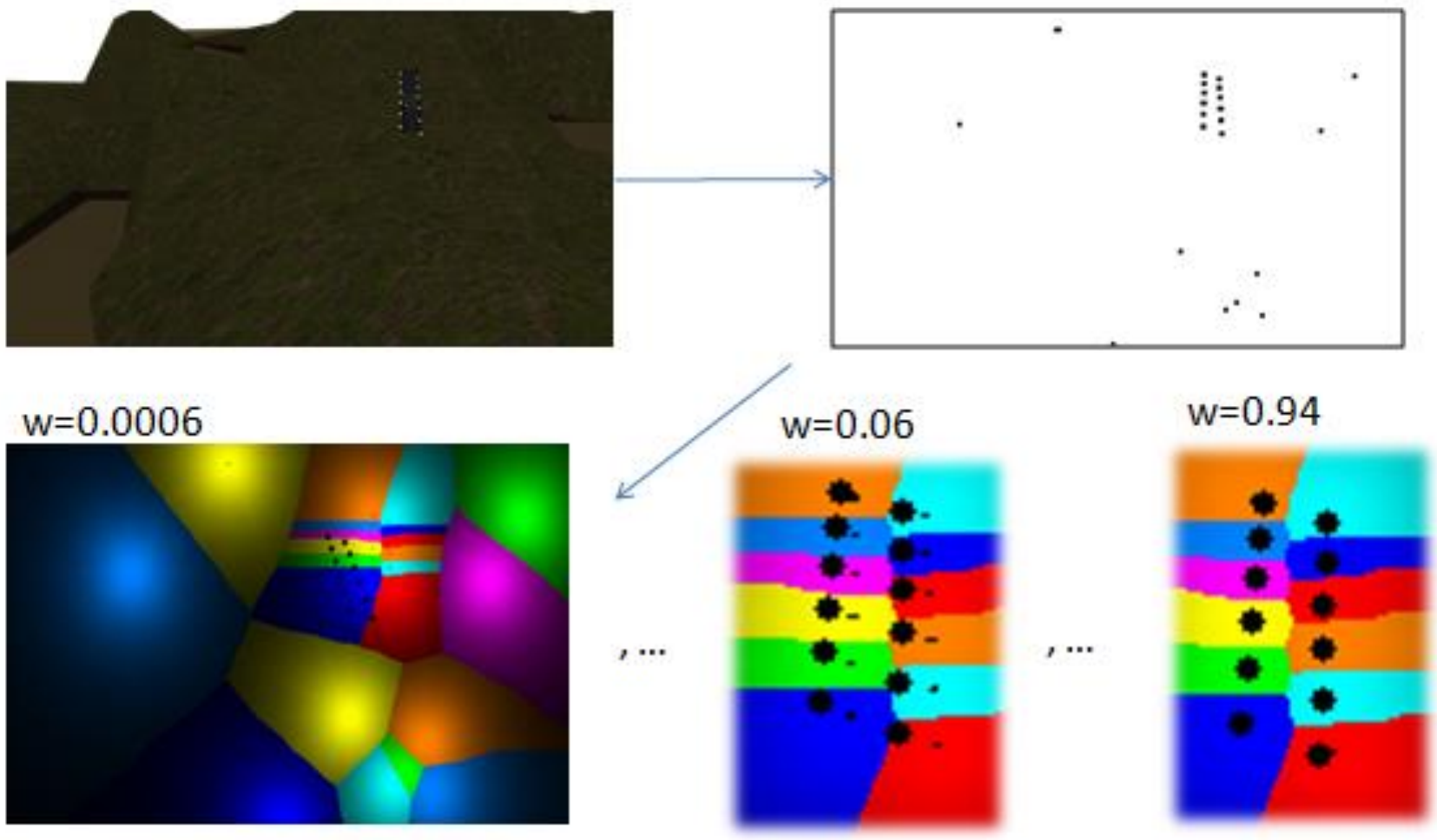

**Fig. 4**. Beacons and some outliers are detected on synthesized image, distance transform is applied and multiple hypothesis are matched. *Black circles*: projections of three sets of beacons (each for one hypothesis) to distance transform map (labeling is not used). *Black dots* show detected beacons positions (on last image are hidden below projected circles due to precise match). Matching weight for the cases is denoted by *w*.

## 4 Evaluation

At the moment of paper writing we have recorded real glide trajectory of UAV and it's IMU signals during landing. We model a video input sequence by Blender (fig.5). The goal of evaluation was to compare computed trajectory with real glide trajectory of UAV based on simulated image input and recorded IMU signals.

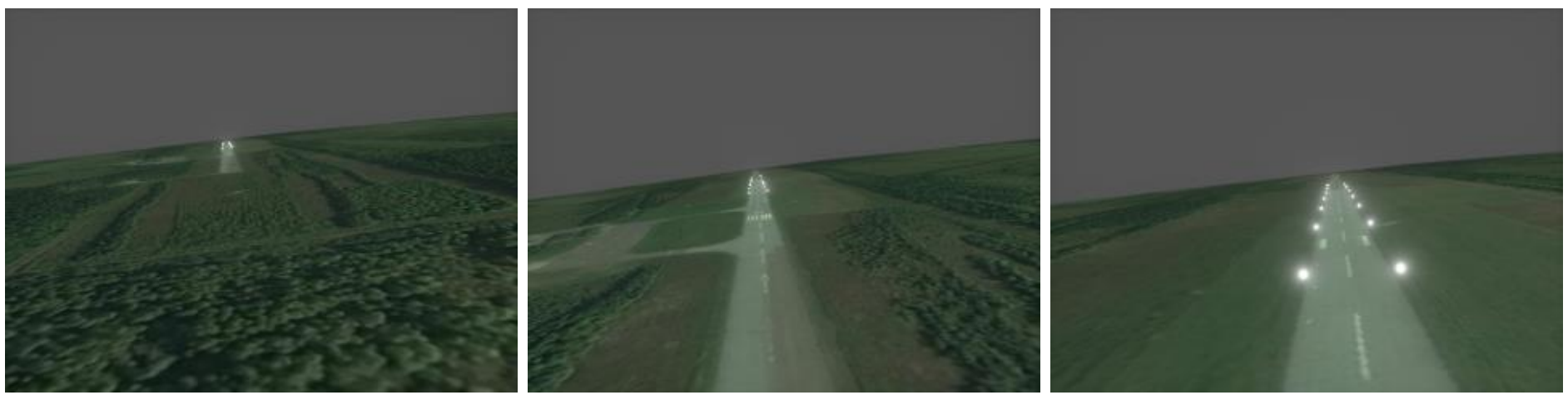

**Fig. 5**. Images from simulated video input.

We use 16 markers and calibrated camera model. Our implementation works online on a personal computer with the x86 processor Pentium (R) Dual-Core CPU with 2.00 GHz. The comparable results are shown on fig. 6 and detailed on table 1.

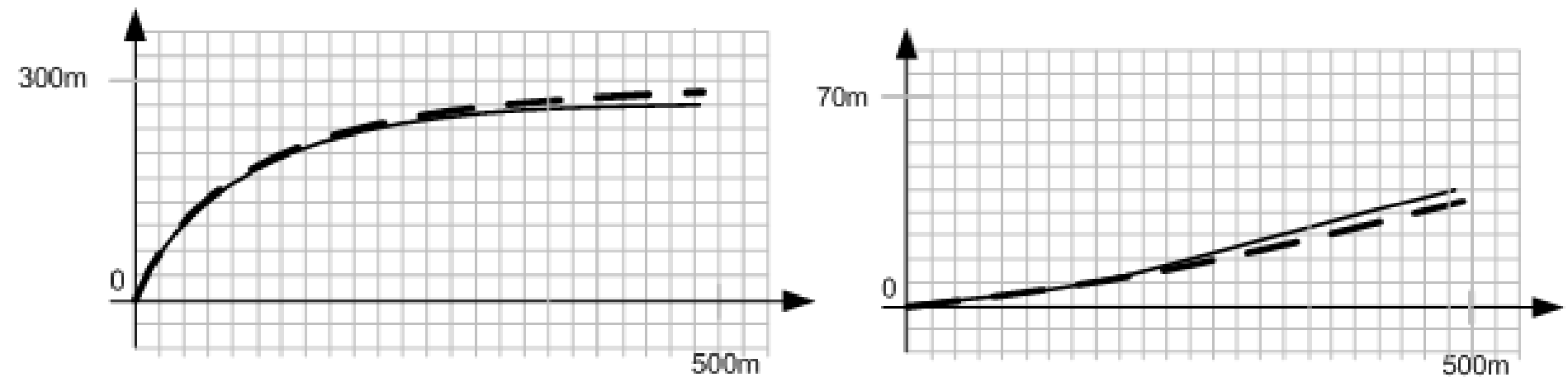

**Fig.6**. Computed (dashed) and recorded glide trajectory. Top view on trajectory (left picture), side view on trajectory (right image)

**Table 1.** Compare recorded and calculated glide trajectory

| Parameter | Value | Units |
|---|---|---|
| Maximum linear deviation across the distance from landing point | 4,12 on 500; 1,52 on 100; 0,15 on 10 | meters |
| Maximum orientation deviation across the distance from landing point | 5,3 on 500; 1,5 on 100; 0,1 on 10 | angles |

The results presented in table 1 show the accuracy of computed trajectory by our algorithm implementation across the distance from landing point (like angular variation).

## 5 Conclusion

Autonomous fixed-wing UAV landing task in GPS-denied environment is one of the actual tasks in UAV navigation problem area. We propose a method of UAV navigation on landing stage based on computer vision, IMU sensor data fusion and infrared dot markers. Described algorithm was evaluated in simulated environment and the comparable results are presented. Infrared dot markers are rather convenient for use on practice.

As a future directions of project development we plan to evaluate our system on real data (like a left image on figure 2, but with MIR markers) and optimize some part of algorithm for onboard using. Also we plan to investigate other markers for autonomous landing, e.g. flashing markers or laser beam markers.

**Acknowledgments.** This work was supported by the Ministry of Education and Science of the Russian Federation (RFMEFI60714X0088) agreement for a grant on 'Development of methods and means of processing and intelligent image analysis and flow of data obtained from a set of stationary and mobile sensors, using high-performance distributed computing for the tasks of monitoring the indoor placement and adjacent outdoor territories'.